\documentclass[letterpaper, 10 pt, conference]{ieeeconf}  % Comment this line out if you need a4paper

\IEEEoverridecommandlockouts                              % This command is only needed if 
\usepackage{cite}
\usepackage{multirow}
\usepackage{graphicx} % DO NOT CHANGE THIS
\usepackage{subfigure}
\usepackage{booktabs}
\usepackage[flushleft]{threeparttable}

\usepackage{pifont} 
\newcommand{\cmark}{\ding{51}}
\newcommand{\xmark}{\ding{55}}

\usepackage{amsmath} % assumes amsmath package installed

\title{\LARGE \bf
Rethinking Dimensionality Reduction in Grid-based 3D Object Detection
}
\author{Dihe Huang$^{1}$, Ying Chen$^{2}$, Yikang Ding$^{1}$, Jinli Liao$^{1}$, Jianlin Liu$^{2}$, \\ Kai Wu$^{2}$, Qiang Nie$^{2}$, Yong Liu$^{2}$, Chengjie Wang$^{2}$, Zhiheng Li$^{1}$
\thanks{$^{1}$Tsinghua University, China, \{\tt\small hdh20, dyk20, liaojl20\}@mails.tsinghua.edu.cn, zhhli@tsinghua.edu.cn}% 
\thanks{$^{2}$Tencent YouTu Lab, China, \{\tt\small mumuychen, jenningsliu, lloydwu, stephennie, choasliu, jasoncjwang\}@tencent.com}
}

\begin{document}

\maketitle
\thispagestyle{empty}
\pagestyle{empty}

%%%%%%%%%%%%%%%%%%%%%%%%%%%%%%%%%%%%%%%%%%%%%%%%%%%%%%%%%%%%%%%%%%%%%%%%%%%%%%%%
% BEV检测的优点，但是已有的方法将点云或3D体素降维到2d bev空间往往会丢失许多信息

% 目前大多数点云目标检测方法都是在Bird eye view 上进行3d目标检测的，因为在bev空间上可以使用2d卷积，应用2d检测head，并且没有遮挡的影响。但现有的方法直接对点云或体素进行降维处理，严重丢失3d空间的信息。因此我们提出xxx，其通过动态高度特征编码来融合不同高度的特征，同时在不同尺度的BEV特征中加入体素残差，使得更好的保留3D空间的信息。
\begin{abstract}
% Most of the current state-of-the-art point cloud object detectors are performed on the bird's eye view (BEV) since methods of 2D detection can be applied directly on BEV space and are not affected by occlusions. 
Bird's eye view (BEV) is widely adopted by most of the current point cloud detectors due to the applicability of well-explored 2D detection techniques. However, existing methods obtain BEV features by simply collapsing voxel or point features along the height dimension, which causes the heavy loss of 3D spatial information. To alleviate the information loss, we propose a novel point cloud detection network based on a Multi-level feature dimensionality reduction strategy, called MDRNet. In MDRNet, the Spatial-aware Dimensionality Reduction~(SDR) is designed to dynamically focus on the valuable parts of the object during voxel-to-BEV feature transformation. Furthermore, the Multi-level Spatial Residuals~(MSR) is proposed to fuse the multi-level spatial information in the BEV feature maps.
% It performs voxel-to-BEV feature transformation by Spatial-aware Dimensionality Reduction (SDR) making it possible to dynamically focus on the parts of the object that are beneficial for the detection task. 
%Furthermore, Multi-level Spatial Residuals~(MSR) is employed to enable BEV feature maps to fuse multi-level spatial information. 
Extensive experiments on nuScenes show that the proposed method outperforms the state-of-the-art methods. Code will be available upon publication.

\end{abstract}

%%%%%%%%%%%%%%%%%%%%%%%%%%%%%%%%%%%%%%%%%%%%%%%%%%%%%%%%%%%%%%%%%%%%%%%%%%%%%%%%
% 阐述现在3D检测的现状，我们的优势

\section{INTRODUCTION}
LiDAR-based 3D object detection is an important task in computer vision which has a wide range of applications in autonomous driving and robotics. One of the major challenges in this task is how to learn the various object properties from the sparse and unstructured point clouds. To solve this problem, many methods~\cite{shi2019pointrcnn, 3dssd} utilize PointNet-like~\cite{qi2017pointnet, qi2017pointnet++} networks to extract 3D object properties. 
Though great improvement has been achieved, such methods introduce heavy computational complexity on point sampling and grouping, which makes them unsuitable for large-scale autonomous driving scenes. 
% Though great improvement has been achieved, such methods spend a lot of time on point sampling and grouping, which makes them unsuitable for large-scale autonomous driving scenes. 

To achieve efficient 3D object detection, most of the existing state-of-the-art methods propose to use the grid-based representation and perform 3D object detection in the 2D BEV space.
% updated by choasliu
% Grid-based methods can be further divided into voxel-based\cite{zhou2018voxelnet, Deng2022, chen2022focal, yan2018second} and pillar-based\cite{lang2019pointpillars, shi2022pillarnet} methods, both of which need to perform a dimensionality reduction step along the height dimension~(Z-axis) to obtain 2D BEV features for object detection. Generally, voxel-based approach first voxelizes the point cloud, then uses an encoder consisting of 3D sparse convolutions to extract voxel features, and finally flatten the feature along Z-axis to obtain the BEV features. In pursuit of higher efficiency, pillar-based approach mainly uses the grids which have equal height to the predefined 3D space in the voxelization process. Specifically, pillar-based methods first use the pooling operation to perform point clouds collapse along the height dimension for obtaining BEV features, which will be then encoded by a 2D convolution network to capture the information for the final detection. 
% % However, existing grid-based methods lose a lot of geometric information during BEV feature extraction because they do not utilize geometric information of each individual object in the feature reduction step.
% Despite the above mentioned frameworks are widely used, they do not consider to exploit the geometric structure of each individual object during the dimensionality reduction process, resulting in a lot of geometric information lost in BEV features.
Grid-based methods can be classified into voxel-based\cite{zhou2018voxelnet, Deng2022, chen2022focal, yan2018second} and pillar-based methods\cite{lang2019pointpillars, wang2020pillar, shi2022pillarnet}, both need to perform a dimensionality reduction step along the height dimension~(Z-axis) for 2D BEV feature generation. 
The voxel-based approach first uses a 3D sparse convolutional encoder to extract features from voxelized point clouds and then flatten along Z-axis to obtain BEV features. 
Similarly, pillar-based methods first use a pooling operation to perform point cloud collapse along the height dimension to attain BEV features, followed by a 2D convolution network for final detection. 
Generally, the grid-based methods squeeze the height dimension to achieve high efficiency in the 3D object detection.
Although frameworks mentioned above are predominant, the geometric information loss in the dimensionality reduction process along the Z-axis is seldom discussed, especially for categories with diverse sizes and structures that need adaptive feature extraction.

As a result, previous grid-based methods perform poorly on objects with complex and diverse geometry, such as the motorcycle and bicycle categories. This phenomenon is caused by two issues. First is the use of fixed pooling operations and static convolution kernels in the dimensionality reduction from 3D to BEV. For objects with different structures, recognizing objects in the BEV space becomes more difficult if the dimensionality reduction step cannot adaptively preserve spatial information according to the objects themselves. The second problem is that by downsampling the voxels, the network concentrates on hierarchical semantic information but loses the sparse geometric information which is essential for recognition and localization.
%The second problem results from downsampling the voxels. The network  hierarchical semantic information but loses the sparse geometric information, which is essential for recognition and localization.

% 5. Contribution
In this work, we propose a novel backbone network for grid-based 3D object detection, MDRNet, which is capable of adaptively integrating 3D geometric information into the network. The network is a dual-branch structure, consisting of a lightweight voxel branch and a BEV branch. We introduce two new modules to enhance the retention of 3D geometric information in BEV features. The first is {\em Spatial-aware Dimensionality Reduction}~(SDR). SDR focuses on important features by estimating spatial distribution and performs dynamic aggregation of voxel features along the height dimension.
% It estimates {\em spatial distribution} for the dynamic aggregation of voxel features along height dimension. 
The second is {\em Multi-level Spatial Residuals}~(MSR), which organically fuses voxel features and BEV features at each stage to enrich the multi-level 3D geometric information of the BEV branch. Specifically, the initial features of the BEV branch are obtained from the voxel features via SDR, and at each subsequent stage, the voxel features are fused with the BEV features via MSR, as shown in Fig.~\ref{overview}(c). 
% Considering the reasons mentioned above, a very intuitive idea is to perform dimensionality reduction operations adaptively based on the geometric information of the object. In addition, we propose a novel network design concept to aggregate the spatial features of different receptive fields into BEV space. We name the proposed operation as Spatial-aware Dimensionality Reduction~(SDR) and the proposed network structure as Multi-level Spatial Residuals~(MSR). 

The proposed backbone can be an easy replacement for the backbone network of existing grid-based point cloud object detectors. To verify the effectiveness, we apply our method on existing 3D object detection frameworks~\cite{shi2020pv, Yin2020centerpoint}. Our approach achieves a dramatic enhancement on both the KITTI~\cite{geiger2013kitti} and nuScenes~\cite{caesar2020nuscenes} datasets with almost the same runtime. 
% Without bells and whistles, our method exceeds SOTA methods in the nuScenes {\em test} set.
To summarize, we provide the critical insights into the proposed 3D detection method:
\begin{itemize}

% ===== first version ======
% \item We introduce a voxel-to-BEV feature transformation operation named Spatial-aware Dimensionality Reduc-tion (SDR), which uses the estimated spatial distribution as weights for dynamic feature aggregation along the Z-axis. 
% % The purpose of the learned spatial distribution is to focus on the easily detectable geometric structure of the objects, thus preserving the 3D information well during the dimensionality reduction.
% % which weighted the voxel for different objects to predict the BEV feature.

% % \item We propose a voxel-residual connection for 3D voxel feature and 2D feature fusion that allows multi-level feature fusion to guarantee the establishment of geometry information protection.
% \item We propose a novel network design idea with multi-level spatial-residual connections (MSR), which allows multi-scale 3D-BEV connections to guarantee the establishment of geometric information protection.

% ==== second version =====
\item We design a universal backbone, named MDRNet, which is readily used with any grid-based point cloud detectors to enrich features extracted from dimensionality reduction. The backbone estimated spatial distribution for dynamic feature aggregation and enabled multi-level 3D-BEV connections without incurring additional computational burden. 

\item We propose two novel modules: Spatial-aware Dimensionality Reduction (SDR) and Multi-level spatial-residual connections (MSR). With the combination of two modules, geometry information of point clouds can be successfully retained and aggregated in the dimensionality reduction process, which significantly boosts the 3D detection performance.

\item Extensive experiments and analyses demonstrate that MDRNet outperforms various strong baselines and achieves SOTA results on point cloud object detection task.

\end{itemize}

\section{RELATED WORK}
Relevant prior grid-based representation 3D detection work includes studies of voxel-based and pillar-based for feature representation.

\subsection{Voxel-based Methods}
VoxelNet\cite{zhou2018voxelnet} conducted one of the first studies of end-to-end 3D detection that divides the irregular point cloud into voxels and predicts accurate 3D bounding boxes. While VoxelNet, each convolutional middle layer applies 3D convolution with high computational cost, which makes it challenging to use for real-time applications. SECOND~\cite{yan2018second} introduces 3D sparse convolution for acceleration and performance improvement. It extracts voxel features using a backbone network composed of 3D sparse convolution, and then concatenates the voxel features along the height dimension, followed by 2D convolution layers to obtain dense BEV features.
% Methods SECOND\cite{yan2018second} and FocalsConv\cite{chen2022focal} seek sparse spatially convolutional for efficient inference. 
On the basis of SECOND~\cite{yan2018second}, CenterPoint~\cite{Yin2020centerpoint} proposes to use a single positive cell for each object. VISTA\cite{Deng2022} project the 3D feature maps into the bird’s eye view(BEV) and range view(RV), performing multi-view transformer fusion. FocalsConv~\cite{chen2022focal} proposes a dynamic mechanism, aiming to make the learning process focused on the foreground data. 
% Besides, LargeKernel3D~\cite{chen2022scaling} remains a spatially large kernel size to enlarge the receptive fields and model capacity but shares weights among local neighbors to keep efficiency. 
However, most of the existing methods~\cite{Yin2020centerpoint, deng2021voxel, mao2021votr, Deng2022, chen2022focal} are based on the framework of SECOND~\cite{yan2018second}, which does not take advantage of the multi-scale geometric information. In contrast, our proposed method retains multi-scale geometric information onto the BEV features used for object detection.
% In this work, we use a centerpoint head with sparse convolutional method. 

%Recently SECOND\cite{yan2018second} improved the inference speed of VoxelNet by the spatially sparse convolutional network, but the 3D convolutions remain a bottleneck. In contrast to regular sparse convolution which dilates all input features to its kernel-size neighbors, FocalsConv\cite{chen2022focal} dynamically determines which input features deserve dilation and dynamic output shapes using predicted cubic importance. Different 

\subsection{Pillar-based Methods}
% Voxel-based~\cite{yan2018second, chen2022focal} methods have done much work in speeding up and saving memory. Still, as the resolution of 3D voxel increases, it isn't easy to balance its computational complexity and memory. 
Compared with the voxel-based approach, the pillar-based approach~\cite{lang2019pointpillars, wang2020pillar, shi2022pillarnet} aims to reduce the time consumption during inference. PointPillars~\cite{lang2019pointpillars} utilizes PointNets~\cite{qi2017pointnet} to encode point features and then uses a pooling operation to transform the point features into a pseudo-image in bird's eye view, enabling end-to-end learning with only 2D convolutional layers. Thus PointPillars~\cite{lang2019pointpillars} can be deployed on embedded systems with low latency and computation consumption.
Infofocus~\cite{wang2020infofocus} adds a second-stage attention network to PointPillars~\cite{lang2019pointpillars} for fine-grained proposal refinement.
PillarNet~\cite{shi2022pillarnet}, a modified version of CenterPoint-pillar~\cite{Yin2020centerpoint}, introduces the 2D sparse convolution of ResNet18 structure into the backbone for BEV feature extraction. Experiments show that after sufficient 2D sparse convolution extraction, the pillar-based network can achieve the similar accuracy as the voxel-based. However, due to the large amount of 3D geometric information lost, the pillar-based methods are difficult to break the bottleneck of 3D object detection.

\section{METHOD}
% 我们的目标是融合pillar-based和voxel-based表征方法，尽可能保留对应的3D信息，除此外，不同object的属性依赖于不同高度的特征，所以我们对特征量化也做了相应处理。
\begin{figure*}[ht]
    \centering
    \includegraphics[width=0.97\textwidth]{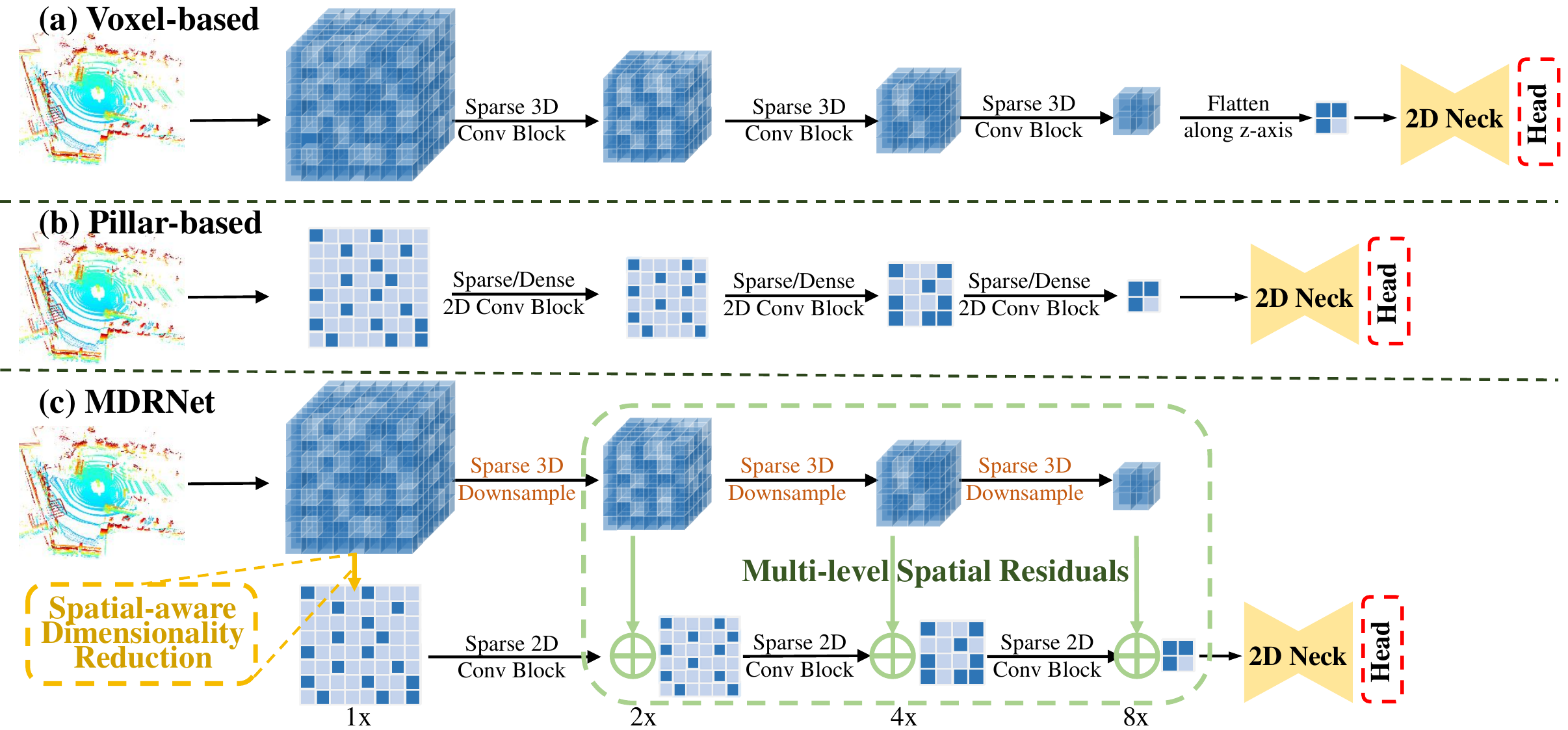}
    \caption{Overview of different 3D feature encoding architectures. The first and second rows are the voxel-based and pillar-based~(BEV-based) paradigms, respectively. The last row is the proposed \textbf{MDRNet}, which consists of voxel branches and BEV branches. In each stage, voxel features are connected to BEV features by an elaborate dimensionality reduction operation, which enables the BEV features to retain multi-scale 3D geometric information.}
    \label{overview}
\end{figure*}
In this section, we first review the previous feature reduction operations in~\ref{preliminaries}. Then, we describe our proposed Spatial-aware Dimensionality Reduction (SDR) and Multi-level Spatial Residuals (MSR) in Sec.\ref{sec_sdr} and Sec.\ref{sec_msr}, respectively. Finally, we present our overall network structure in Sec.\ref{sec_network}. For simplifying, we denote the Z-axis as the height dimension

%%%%%%%%%%%%%%%%%%%%%%%%%%%%%%%%%%%%%%%%%%%%%%%%%%%%%%%%%%%%%%%%%%%%%%%%%%%%%%%%
\subsection{Preliminaries} 
\label{preliminaries}
% 现有的基于voxel 或 pillar的方法沿着高度维度使用池化或拼接的形式将三维的特征降维到2维的BEV特征
% 但是3D目标检测 自然地focus on 目标的有区分性的重要特征。
% 对于3D目标检测中的任一个object而言，由于点云的不规则结构以及物体不同高度的特征保留着重要程度不一的3d信息，在降维时候应当对重要位置的特征给予较高的权重，而对不重要的特征给予较低的权重，均值池化对每个高度的特征等价处理，因此会丢失3D信息 而 最大值池化只保留部分特征因而也会丢失3D信息。
% 对于3D目标检测中的不同object而言，由于不同object的几何结构相差较大, 不同object进行维度坍塌过程中同一高度信息的权重应当不同的， 因此沿着高度对特征进行拼接再使用2d卷积减少特征维度是会导致3d信息的丢失。

\begin{figure}[ht]
\centering
\subfigure[Max/Mean pooling]{
	\begin{minipage}[t]{0.22\textwidth}
		\includegraphics[width=1\textwidth]{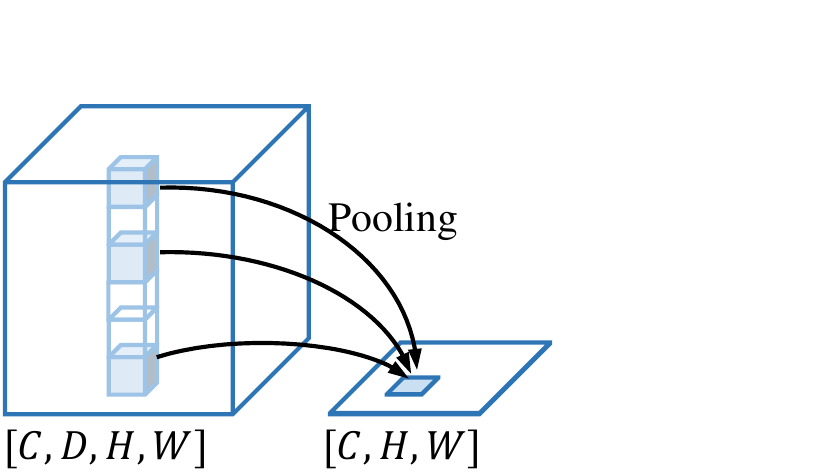}
	\end{minipage}
    \label{fig:pillar}
} 
\subfigure[Flatten along z-axis and reduce features]{
	\begin{minipage}[t]{0.22\textwidth}
  	\includegraphics[width=1\textwidth]{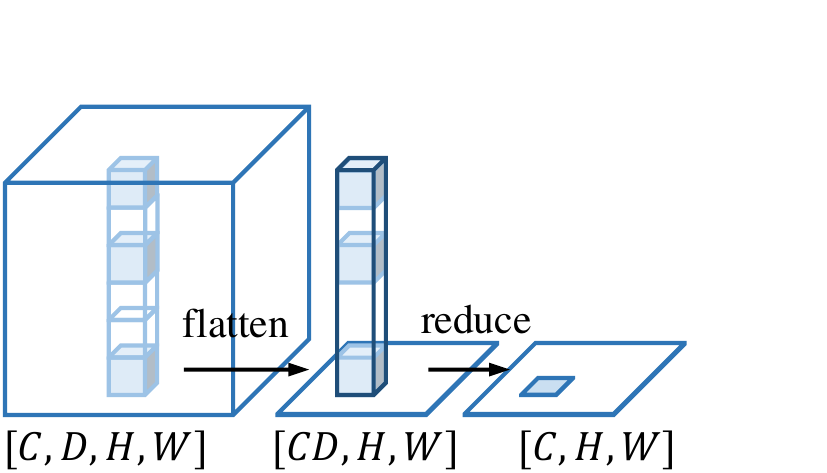}
	\end{minipage}
    \label{fig:voxel}
}\\
% \subfigure[Convolution with kernel size and stride size both $D$]{
\subfigure[3D Sparse Convolution]{
	\begin{minipage}[t]{0.22\textwidth}
  	\includegraphics[width=1\textwidth]{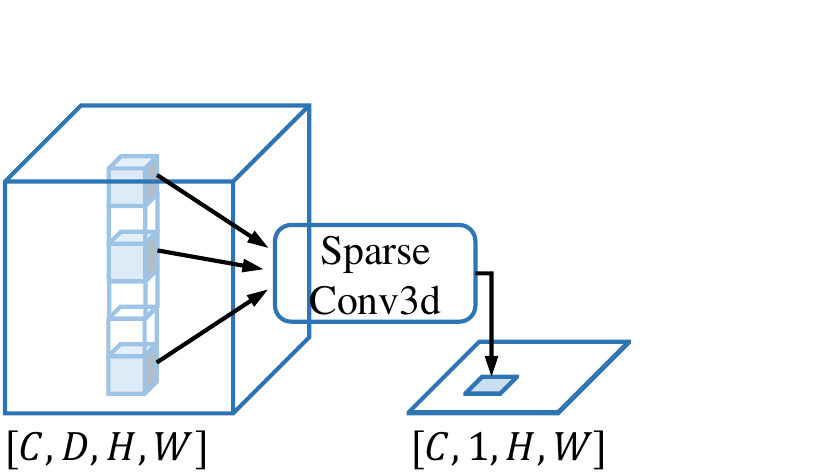}
	\end{minipage}
\label{fig:conv}
}
\subfigure[SDR]{
	\begin{minipage}[t]{0.22\textwidth}
  	\includegraphics[width=1\textwidth]{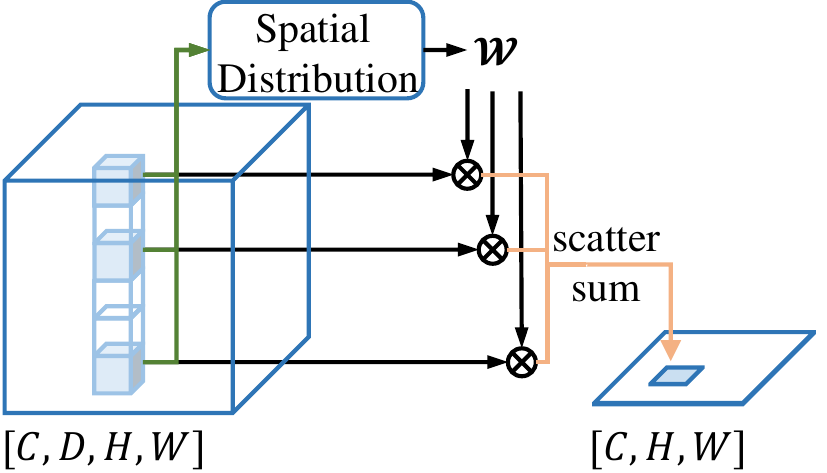}
	\end{minipage}
\label{fig:sdr}
}
\caption{Different approaches of dimensionality reduction.}
\label{quant}
\end{figure}

% 本文中我们提出一种用于3d目标检测的新颖的点云特征提取网络，其能够有效的将3d空间信息保留在2D的bev 特征图上，
On the BEV representation of point cloud, there is no need to consider object occlusion and the efficient 2D convolution can be leveraged. Existing voxel- or pillar-based methods project 3D features to bird's eye view through dimensionality reduction. Pillar-based methods~\cite{shi2022pillarnet, lang2019pointpillars} directly pool the point cloud along the Z-axis to obtain the BEV representation of point cloud,
%allowing subsequent BEV feature encoding using 2d convolution,
as shown in Fig.\ref{fig:pillar}. The voxel-based methods mainly use the SECOND~\cite{yan2018second} architecture to extract 3D features and concatenate these features along the z-dimension, followed by 2d convolutions to reduce feature dimensionality, as shown in Fig.\ref{fig:voxel}. Obviously, Fig.\ref{fig:voxel} computes many empty voxels, which can be efficiently replaced by the sparse convolution with kernel size equal to the z-dimensional range, as in Fig.\ref{fig:conv}. Given an input feature $\mathrm{x}_{i, j, k}$ in the spatial space, the above  dimensionality reduction process can be expressed uniformly as
\begin{equation}
    \mathrm{y}_{i,j} = \sum_{k\in \mathrm{Z}_{i,j}} \mathrm{w}_{i,j,k} \cdot \mathrm{x}_{i,j,k},
\label{eq1}
\end{equation}
where $i, j, k$ are the coordinates along the $\mathrm{X, Y, Z}$ axes, $\mathrm{Z}_{i,j}$ is the number of features whose coordinates of $x, y$-axis are equal to $<i,j>$, $\mathrm{w}_{i,j,k}$ is the weight of each feature and $\mathrm{y}_{i,j}$ is the output BEV feature at position $<i,j>$. When using mean pooling for dimensionality reduction, $\mathrm{w}_{i,j,k}$ equals $1/\mathrm{Z}_{i,j}$. When using max pooling, $\mathrm{w}_{i,j,k}$ is a binary weight that is $1$ only when $\mathrm{x}_{i,j,k}$ is the maximum the along z-axis. When using convolution, $\mathrm{w}_{i,j,k}$ is is a static learnable parameter optimized with the training process and its value is fixed once the training is over. 

All the three dimensionality reduction operations will cause the loss of spatial information. (\romannumeral1) Mean pooling simply averages the spatial features at the same $<i,j>$ position without using the semantic information.
%to determine the weights of features.
(\romannumeral2) Max pooling only retains the maximum value of features along the z-axis, leading to loss of a lot of relevant information. (\romannumeral3) Convolution operation assigns different weights to the features along the Z-axis, but the weights are identical for different objects. Due to the large variation in geometric structure of different objects, using fixed weights without adaptive adjustment according to the object geometric information would obscure the valuable information and limit the representation capability for point cloud data. 
% , further diminishing the object perception capability of the 3D object detector.
Since 3D object detection should focus on different locations of different instances, feature encoding using the same weights for different instances along the Z-axis will further reduce the object perception capability of the 3D object detector.
 
%%%%%%%%%%%%%%%%%%%%%%%%%%%%%%%%%%%%%%%%%%%%%%%%%%%%%%%%%%%%%%%%%%%%%%%%%%%%%%%%
\subsection{Spatial-aware Dimensionality Reduction (SDR)} 
\label{sec_sdr}
% For an object, due to the irregular structure of the point cloud and the complex geometric structure of the object, 3D object detection should naturally pay different attention to the features at different height position. For the whole point cloud scene, due to the large variation in geometric structure of different instances, the weights along the height dimension should be impacted dynamically by the semantics of the object.
Considering categories like pedestrians, motorcycles and bicycles, has more complex geometries, we argue that not all parts of the object are equally important to the task of point cloud object detection. A dynamic focus on the easily detectable parts of the object should be considered.
In this subsection, we present a novel dimensionality reduction structure called Spatial-aware Dimensionality Reduction (SDR) that dynamically focuses on the easily detectable parts of the object and effectively preserves the geometric information. Compared to the previously described approaches, SDR explores the spatial correlation distribution and geometric semantic features along z-axis. Learning spatial distribution enables the network to focus on the easily detectable geometric structure of the objects and preserve the 3D information well during the dimensionality reduction.
% This spatial distribution is learned in order to explore which position along the Z-axis of the object is more favorable for the 3D target detection task.
% and the dimensionality reduction operation is transformed into a dynamic weighted sum of z-axis features with the predicted geometric distribution as the weight. 

The proposed SDR is illustrated in Fig.\ref{fig:sdr}.
We denote $\{\mathrm{w}_{i,j,k}|k\in \mathrm{Z}_{i,j}\}$ in Eq.~\ref{eq1} as the spatial correlation distribution of $(i, j)$ position along the Z-axis, which represents the geometric semantic representation capability of the voxel features. Due to the various geometric structures of objects, $\mathrm{w}_{i,j,k}$ should be dynamically generated according to the geometric semantics of objects. Therefore, we first use a submanifold sparse convolution $\mathcal{F}(\cdot)$ with kernel size 3 to encode the geometric information of the neighborhood of voxel and the output is represented as $\widetilde{w}(\mathrm{x}_{i,j,k}) = \mathcal{F}(U(\mathrm{x}_{i,j,k}))$, where $U$ denotes the set of neighbors within a distance of 3.
Then, a normalization function is used to transform $\widetilde{w}$ into a probability distribution. In the following, we show several forms of normalization functions:
\begin{equation}
\begin{aligned}
    & \mathrm{w}_{i,j,k}^{ReLU} = \frac{ReLU(\widetilde{w}_{i,j,k})}{\sum_{l\in \mathrm{Z}_{i,j}} ReLU(\widetilde{w}_{i,j,l})},\\
    & \mathrm{w}_{i,j,k}^{Sigmoid} = \frac{1}{1+Exp(-\widetilde{w}_{i,j,l})}, \\
    & \mathrm{w}_{i,j,k}^{Softmax} = \frac{Exp(\widetilde{w}_{i,j,k})}{\sum_{l\in \mathrm{Z}_{i,j}} Exp(\widetilde{w}_{i,j,l})}, \\
\end{aligned}
\end{equation}
In our ablation studies, we find that $\mathrm{w}^{Softmax}$ achieves superior performance than others. The proposed SDR is spatial-aware and instance-aware, and fully preserves the spatial geometric properties of the point cloud.

\begin{figure*}
    \centering
    \subfigure[Ground truth]{
	\begin{minipage}[t]{0.28\textwidth}
		\includegraphics[width=1\textwidth]{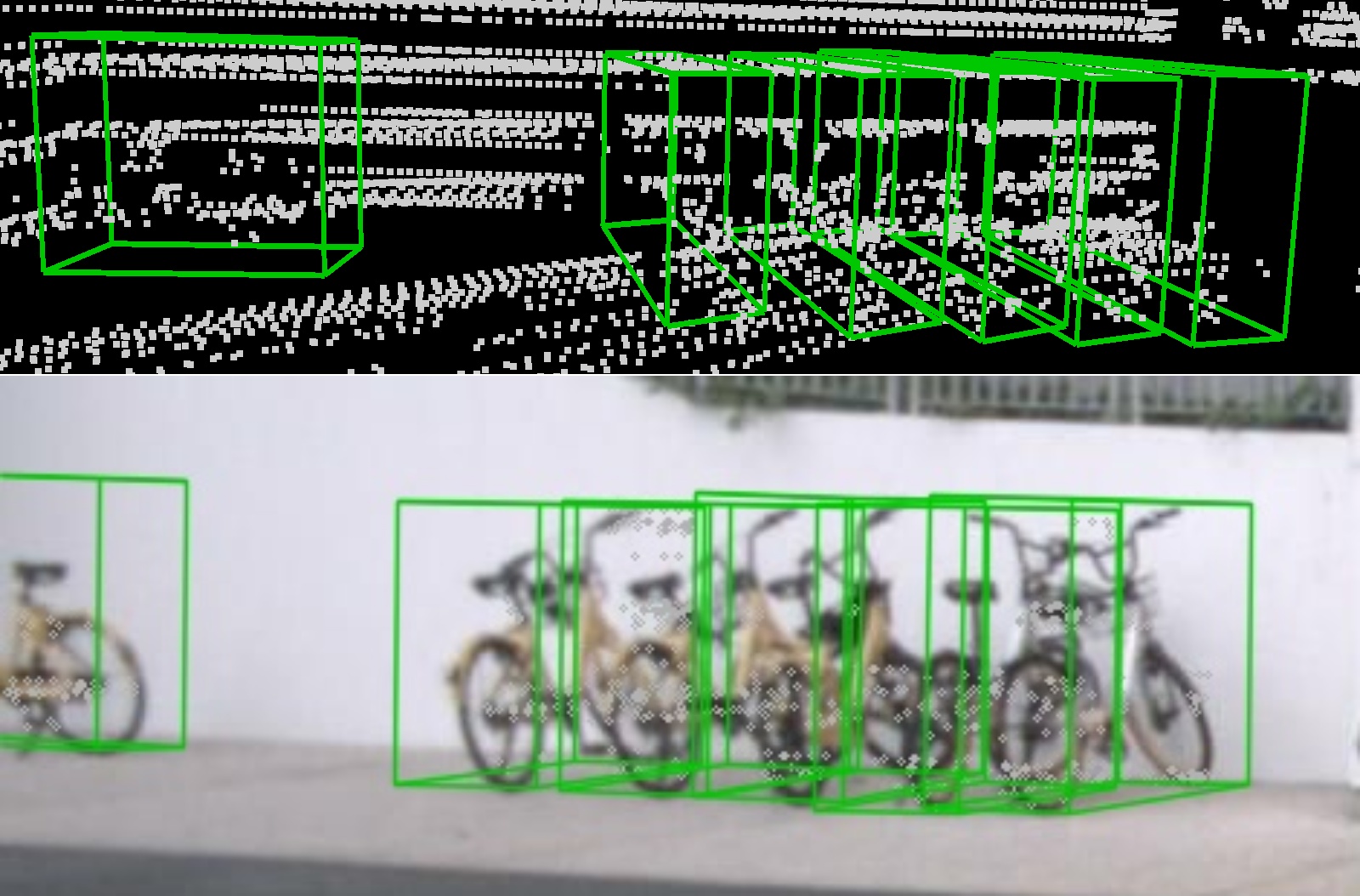}
	\end{minipage}
    \label{viz_gt}
    }
    \subfigure[CenterPoint]{
	\begin{minipage}[t]{0.28\textwidth}
		\includegraphics[width=1\textwidth]{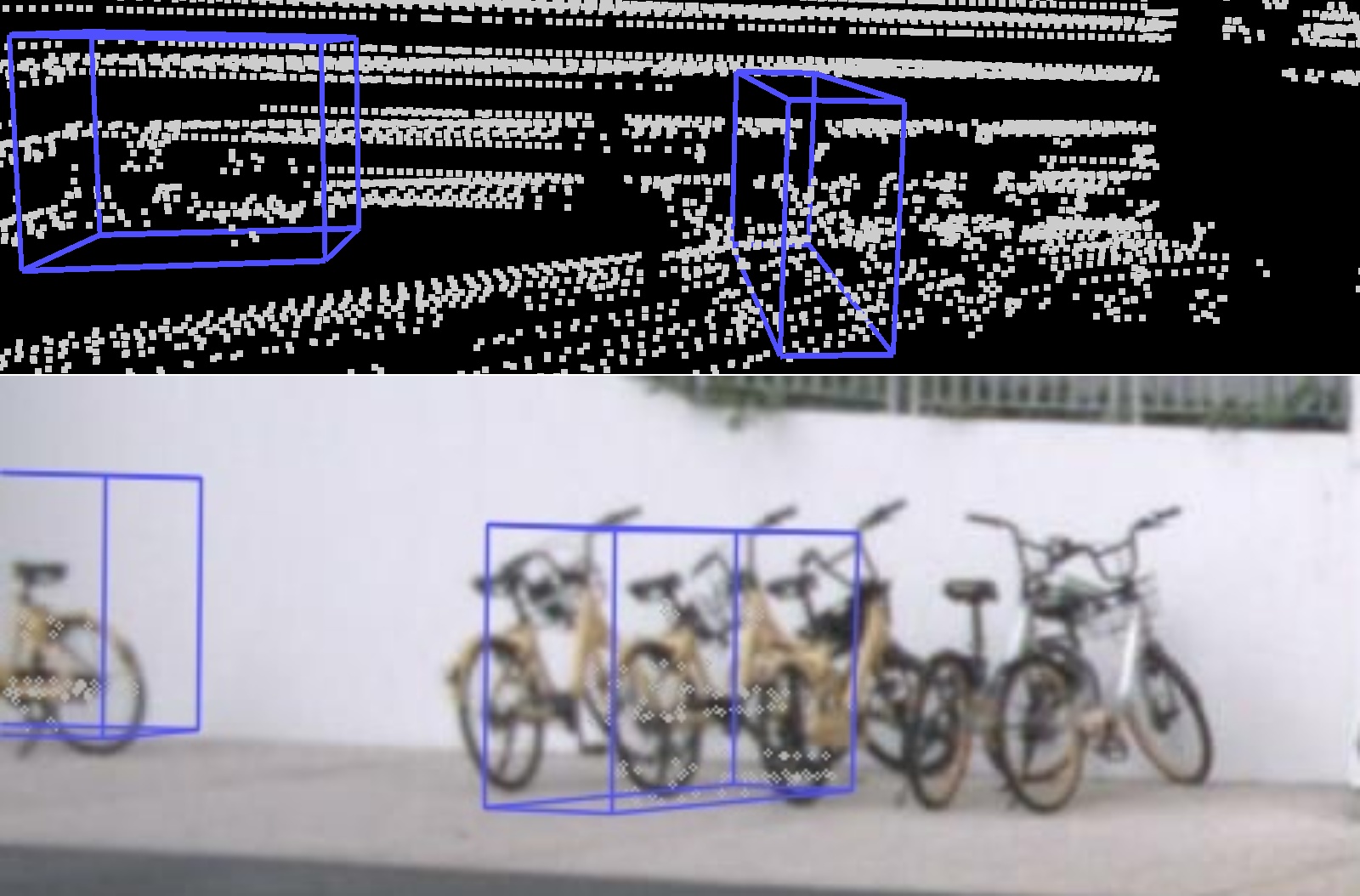}
	\end{minipage}
    \label{viz_centerpoint}
    }
    \subfigure[CenterPoint + MDRNet]{
	\begin{minipage}[t]{0.28\textwidth}
		\includegraphics[width=1\textwidth]{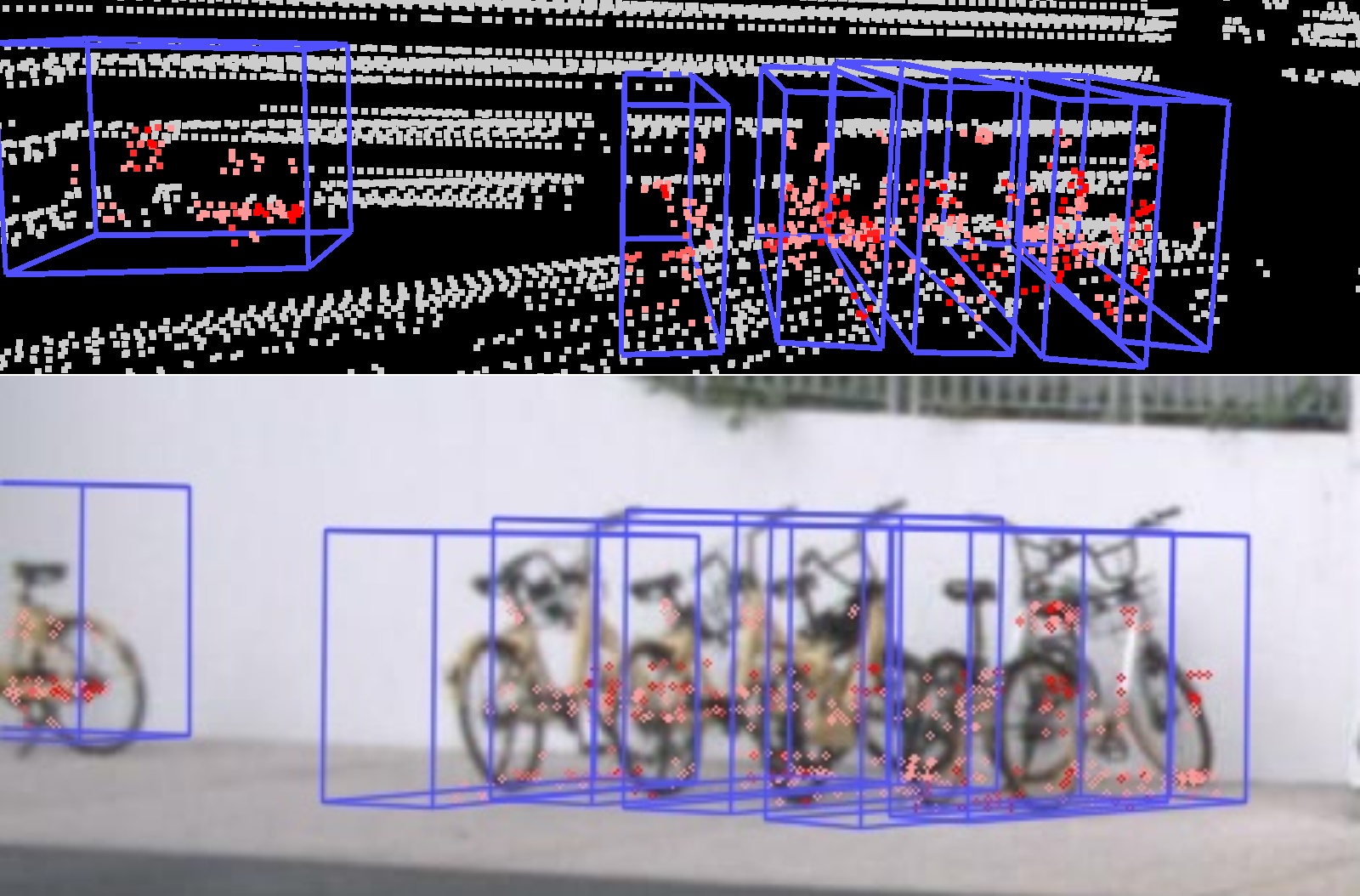}
	\end{minipage}
    \label{viz_mdrnet}
    }
    \subfigure{
	\begin{minipage}[t]{0.04\textwidth}
		\includegraphics[width=0.72\textwidth]{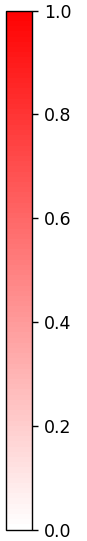}
	\end{minipage}
    }
    \caption{Qualitative results on the nuScenes {\em val} set. The ground-truth bounding boxes are shown in green, and the predicted bounding boxes are in blue. Compared to CenterPoint~\cite{Yin2020centerpoint}, our approach dynamically focuses on the distinguishable regions of objects, making it easier to detect objects with complex geometries, such as bicycles. In the third column, we visualize the spatial distribution learned in the SDR. The closer the color comes to red, the more the network focuses on that area.}
    \label{viz}
\end{figure*}

% Benefiting from the idea of dynamic convolution~\cite{chen2020dynamic}, 
%%%%%%%%%%%%%%%%%%%%%%%%%%%%%%%%%%%%%%%%%%%%%%%%%%%%%%%%%%%%%%%%%%%%%%%%%%%%%%%%
\subsection{Multi-level Spatial Residuals (MSR)}
\label{sec_msr}

% \begin{figure}[ht]
% \centering
% \subfigure[Voxel-based]{
% 	\begin{minipage}[t]{0.1\textwidth}
% 		\includegraphics[width=1\textwidth]{imgs/voxel-based.png}
% 	\end{minipage}
%     \label{fig:voxel_net}
% } 
% \subfigure[Pillar-based]{
% 	\begin{minipage}[t]{0.105\textwidth}
%   	\includegraphics[width=1\textwidth]{imgs/pillar-based.png}
% 	\end{minipage}
%     \label{fig:pillar_net}
% }
% \subfigure[Ours]{
% 	\begin{minipage}[t]{0.2\textwidth}
%   	\includegraphics[width=1\textwidth]{imgs/ours.png}
% 	\end{minipage}
% \label{fig:msr}
% }
% \caption{Overview of different feature representation.}
% \label{overview}
% \end{figure}

With the dimensionality reduction operator, we propose a novel Multi-level Spatial Residuals (MSR) to preserve more geometric information at different resolutions, as shown in Fig.~\ref{overview}(c).
%it is natural to come up with a straightforward idea to preserve more geometric information at different resolutions. we propose a novel backbone network design idea for point cloud object detection, called Multi-level Spatial Residuals (MSR), as shown in Fig.~\ref{overview}(c). 
At each level of resolution, we project the 3D voxel features onto the 2D BEV space and perform element-wise addition to the previous BEV features. In this manner, the 2D feature map is able to retain more 3D geometric information. The cross-dimensional connection between the 3D voxel features of $l+1$ stage and BEV features of $l$ stage is defined by:
\begin{equation}
    \mathrm{y}^{2D}_{l+1} = \mathcal{F}(\mathrm{x}^{3D}_{l}, W) + \mathrm{y}^{2D}_{l},
\end{equation}
where $\mathrm{x}^{3D}$ and $\mathrm{y}^{2D}$ are the 3D voxel features and 2D BEV features. The $\mathcal{F}(\mathrm{x}^{3D}, W)$ can be any 3D to 2D mapping function, such as the previously mentioned max/mean pooling, convolutional dimensionality reduction and our SDR. The design of MSR is able to retain multi-level 3D geometric information of the point cloud to a greater extent. In contrast to pillar-based(Fig.\ref{overview}(b)) and voxel-based~(Fig.\ref{overview}(a)) methods, where the 3D information is retained only at the initial or final resolution, MSR is able to obtain 3D semantics of different sensory fields. The results in Table 1 indicate that MSR can bring a significant improvement to existing 3D detectors without causing additional time consumption.
% Compared to voxel-based and pillar-based methods, 
% We propose a novel backbone network for point cloud object detection, as shown in Fig.\ref{overview}. On 

%%%%%%%%%%%%%%%%%%%%%%%%%%%%%%%%%%%%%%%%%%%%%%%%%%%%%%%%%%%%%%%%%%%%%%%%%%%%%%%%

\subsection{Network Architecture based on SDR and MSR} 
\label{sec_network}
Similar to the previous backbone networks in 3D object detectors~\cite{deng2021voxel, shi2020pv, Yin2020centerpoint, chen2022focal, yan2018second}, our proposed backbone network consists of four stages. 
Different from the voxel-base (Fig.\ref{overview}(a)) and pillar-based (Fig.\ref{overview}(b)) backbone networks, we use the dual-branch network structure, including a BEV branch (pillar branch) and a voxel branch, as shown in Fig.\ref{overview}(c). At the last layer of each stage, we perform element-wise addition to fuse the reduced dimensional geometric features with the BEV feature map. 

Considering the efficiency of the network, each stage of the voxel branch contains only one submanifold sparse convolution with the kernel size 3 and one sparse convolution for down-sampling. The BEV branch consists of residual blocks~\cite{he2016deep} with the number of \{1, 2, 2, 2\} in four stages, respectively. It should be noted that SDR is only used in the first stage and sparse convolution is used in the other stages because the spatial feature spaces in the subsequent stages are downsampled to a limited size and are not sufficient for sparsity adaptation, according to the conclusion of sparsity adaptation in Focals~\cite{chen2022focal}. For the multi-modal version, we simply project the point cloud into image planes to obtain aligned image features extracted from DLA34~\cite{yu2018deep} of pretrained CenterNet~\cite{zhou2019objects, zhou2020tracking}, and then fuse LiDAR and image features by point-wise concatenation before feeding them into the proposed MDRNet. We validate the proposed MDRNet on the existing SOTA 3D detectors~\cite{shi2020pv, Yin2020centerpoint} by directly replacing the backbone network.
% We validate our proposed backbone network in existing SOTA 3D detectors.

\section{Experiments}
% We present ablations and comparisons for MDRNet on the nuScenes and KITTI datasets.
% leader board
\begin{table*}[t!]
\centering
\caption{Performance comparison for 3D object detection on the nuScenes {\em test} set. $\;^{\dagger}$ indicates the flipping test is used. CenterPoint~v2$^{\;\star}$ includes PointPainting~\cite{pointpainting} with pre-trained Cascade R-CNN~\cite{cai2018cascade} and model ensembling}
\label{nus_test}
% \begin{threeparttable}
\resizebox{\textwidth}{!}{
\begin{tabular}{l|c|cc|cccccccccc}
\hline
{\em Method} & {\em Fusion} & NDS & mAP  & Car & Truck & Bus & Trailer & C.V. & Ped & Mot & Byc & T.C. & Bar \\ \hline \hline
PointPillars~\cite{lang2019pointpillars} & \multirow{12}{*}{\xmark} & 45.3 & 30.5 & 68.4 & 23.0 & 28.2 & 23.4 & 4.1 & 59.7 & 27.4 & 1.1 & 30.8 & 38.9 \\ 
3DSSD~\cite{3dssd}           &   & 56.4 & 42.6 &  81.2 & 47.2 & 61.4 & 30.5 & 12.6 & 70.2 & 36.0 & 8.6 & 31.1 & 47.9 \\ 
CBGS~\cite{cbgs}             &   & 63.3 & 52.8 & 81.1 & 48.5 & 54.9 & 42.9 & 10.5 & 80.1 & 51.5 & 22.3 & 70.9 & 65.7 \\ 
HotSpotNet~\cite{hotspotnet} &   & 66.0 & 59.3 & 83.1 & 50.9 & 56.4 & 53.3 & 23.0 & 81.3 & 63.5 & 36.6 & 73.0 & 71.6 \\
CVCNET~\cite{chen2020every}         &   & 66.6  & 58.2 & 82.6 & 49.5 & 59.4 & 51.1 & 16.2 & 83.0 & 61.8 & 38.8 & 69.7 & 69.7 \\ 
CenterPoint~\cite{Yin2020centerpoint}  &   & 65.5 & 58.0 & 84.6 & 51.0 & 60.2 & 53.2 & 17.5 & 83.4 & 53.7 & 28.7 & 76.7 & 70.9 \\ 
CenterPoint$^{\dagger}$         &   & 67.3 & 60.3 & 85.2 & 53.5 & 63.6 & 56.0 & 20.0 & 84.6 & 59.5 & 30.7 & 78.4 & 71.1 \\ 
PillarNet-vgg$^{\dagger}$~\cite{shi2022pillarnet} &   & 69.6 & 63.3 & 86.9 & 56.0 & 62.2 & 62.0 & 28.6 & 86.3 & 62.6 & 33.5 & 79.6 & 75.6 \\
PillarNet-18$^{\dagger}$~\cite{shi2022pillarnet}  &   & 70.8 & 65.0 & 87.4 & 56.7 & 60.9 & 61.8 & \textbf{30.4} & 87.2 & 67.4 & 40.3 & 82.1 & 76.0 \\
PillarNet-34$^{\dagger}$~\cite{shi2022pillarnet}  &   & 71.4 & 66.0 & \textbf{87.6} & 57.5 & 63.6 & 63.1 & 27.9 & 87.3 & 70.1 & 42.3 & 83.3 & 77.2 \\ 
VISTA-OHS$^{\dagger}$~\cite{Deng2022} &   & 69.8 & 63.0 & 84.4 & 55.1 & 25.1 & \textbf{63.7} & 54.2 & 71.4 & 70.0 & 45.4 & 82.8 & \textbf{78.5} \\
Focals Conv~\cite{chen2022focal}  &   & 70.0 & 63.8 & 86.7 & 56.3 & \textbf{67.7} & 59.5 & 23.8 & 87.5 & 64.5 & 36.3 & 81.4 & 74.1 \\ \hline
Ours             & \multirow{2}{*}{\xmark} & 70.5 & 65.2 & 86.5 & 54.5 & 63.8 & 58.9 & 25.7 & 86.6 & 73.1 & 45.2 & 82.9 & 74.8 \\
Ours$^{\dagger}$ &  & \textbf{72.0} & \textbf{67.2} & 87.3 & \textbf{57.7} & 66.5 & 62.2 & 28.3 & \textbf{87.6} & \textbf{74.4} & \textbf{48.5} & \textbf{84.3} & 75.2 \\ 
\hline \hline
PointPainting~\cite{pointpainting}     & \multirow{8}{*}{\cmark} & 58.1 & 46.4 & 77.9 & 35.8 & 36.2 & 37.3 & 15.8 & 73.3 & 41.5 & 24.1 & 62.4 & 60.2 \\ 
3DCVF~\cite{3dcvf}                     &  & 62.3 & 52.7 & 83.0 & 45.0 & 48.8 & 49.6 & 15.9 & 74.2 & 51.2 & 30.4 & 62.9 & 65.9 \\ 
FusionPainting~\cite{xu2021fusionpainting}   &  & 70.4 & 66.3 & 86.3 & 58.5 & 66.8 & 59.4 & 27.7 & 87.5 & 71.2 & 51.7 & 84.2 & 70.2 \\ 
MVP~\cite{yin2021multimodal}                         &  & 70.5 & 66.4 & 86.8 & 58.5 & 67.4 & 57.3 & 26.1 & 89.1 & 70.0 & 49.3 & 85.0 & 74.8 \\
PointAugmenting~\cite{wang2021pointaugmenting} &  & 71.0 & 66.8 & 87.5 & 57.3 & 65.2 & 60.7 & 28.0 & 87.9 & 74.3 & 50.9 & 83.6 & 72.6 \\
CenterPoint v2$^{\star}$        &  & 71.4 & 67.1 & 87.0 & 57.3 & 69.3 & 60.4 & 28.8 & \textbf{90.4} & 71.3 & 49.0 & \textbf{86.8} & 71.0 \\
Focals Conv-F~\cite{chen2022focal}                   &  & 71.8 & 67.8 & 86.5 & 57.5 & \textbf{68.7} & 60.6 & 31.2 & 87.3 & 76.4 & 52.5 & 84.6 & 72.3 \\ 
Focals Conv-F$^{\dagger}$~\cite{chen2022focal}      &  & 72.8 & 68.9 & 86.9 & 59.3 & \textbf{68.7} & 62.5 & 32.8 & 87.8 & \textbf{78.5} & 53.9 & 85.5 & 72.8 \\ \hline
Ours-F & \multirow{2}{*}{\cmark} & 72.1 & 68.1 & 87.5 & 	56.7 & 	65.3 & 	63.2 &	31.1 &	88.5 &	75.3 &	52.0 & 	84.1 &	76.9 \\
Ours-F$^{\dagger}$  &  & \textbf{73.5} & \textbf{69.8} & \textbf{88.1} & \textbf{59.5} & \textbf{68.7} &	\textbf{65.0} & \textbf{32.6}	& 89.0 & 77.4 &	\textbf{55.3} & 84.9 & \textbf{77.1} \\  \hline 

\end{tabular}}
% \end{threeparttable}
\end{table*}

\subsection{Dataset and Technical Details}
\textbf{nuScenes Dataset.} The nuScenes~\cite{caesar2020nuscenes} dataset is a large-scale autonomous driving dataset for 3D perception tasks, which is collect by a 32-bean synced LIDAR, 5 radars and 6 cameras with full 360$^{\mathrm{o}}$ coverage around. It contains 1,000 driving sequences, of which 700 are for training, 150 for validation and 150 for testing. The 3D bounding box annotations of nuScenes detection task include 10 object categories with a long-tailed distribution. For the evaluation, the official metrics are the mean Average Precision~(mAP) and nuScenes detection score~(NDS). Following previous work, 10 LiDAR scans are accumulated as network input and the results are reported using the official evaluation protocol.
% We follow the regular practice of accumulating 10 LiDAR scans as network input and report the results using the official evaluation protocol. 

\textbf{KITTI Dataset.} 
The KITTI~\cite{geiger2013kitti} dataset contains 7,481 training samples and 7,518 testing samples. Following previous works\cite{chen20153d, yan2018second, shi2020pv, chen2022focal},
% We follow the same approach as in~\cite{chen20153d, yan2018second, shi2020pv, chen2022focal} to 
we split the training samples into a {\em train} set of 3712 samples and a {\em validation} set of 3769 training samples. The annotations include three categories~({\em car}, {\em pedestrian} and  {\em cyclist}) which are divided into three difficulty levels~(Easy, Moderate and Hard).
% In the KITTI benchmark, all categories are divided into three difficulty levels: "Easy". "Moderate" and "Hard". 
We evaluate models on {\em val.} set using 3D Average Precision~(AP$_{\textrm{3D}}$) metric, which is calculated with recall 40 positions~(R40). The performance of models is ranked based on the Moderate difficulty samples.
% Models are commonly evaluated in terms of the mean Average Precision~(mAP) metric.

\textbf{Implementation Details.} Our implementation is based on the published code of~\cite{Yin2020centerpoint, shi2022pillarnet}, as well as on the open-sourced OpenPCDet~\cite{shi2020pv, chen2022focal}. The training schedules are the same as the previous works~\cite{Yin2020centerpoint, chen2022focal, shi2022pillarnet}. 

For nuScenes dataset, models are trained with a batch size of 16 for 20 epochs on 4 V100 GPUs. The Adam optimizer is adopted with one-cycle learning rate policy and the momentum range from 0.85 to 0.95. The max learning rate is equal to 1e-3 and the weight decay is set to 0.01. Following the conventional settings, the Z-axis detection range is set as [-5m, 3m]. The X-axis and Y-axis detection ranges are set as [-51.2m, 51.2m] and [-54m, 54m] when the voxel size is (0.1m, 0.1m, 0.2m) and (0.075m, 0.075m, 0.2m), respectively.
% and the X-axis and Y-axis detection ranges are set as [-54m, 54m] when the voxel size is (0.075m, 0.075m, 0.2m).

For KITTI dataset, all networks are trained with a batch size of 16 for 80 epochs. The Adam optimizer is adopted with weight decay 0.01 and momentum 0.9. The learning rate is set as 0.01 and decreases using the cosine annealing strategy. The point cloud range of X, Y and Z axis are clipped to [0m, 70.4m], [-40m, 40m] and [-3m, 1m] respectively. The initial voxel size is equal to (0.05m, 0.05m, 0.1m). 

Following previous methods~\cite{Yin2020centerpoint, chen2022focal, shi2022pillarnet}, data augmentations including random flipping, global scaling, global rotation and ground-truth sampling~\cite{yan2018second} are used to boost the performance of the 3D object detectors. For the ground-truth sampling in multi-modal setting, as with~\cite{wang2021pointaugmenting, chen2022focal}, we copy the corresponding 2D objects in bounding boxes onto images based on the objects' center distance. For the models used to submit results to the nuScenes test server, ground-truth sampling is disabled in the last 4 epochs, as done by \cite{wang2021pointaugmenting, chen2022focal, chen2022scaling}.

\subsection{Overall Results}
% \textbf{nuScenes.} 
We evaluate MDRNet and the multi-modal variant upon CenterPoint~\cite{Yin2020centerpoint} on the nuScenes test server and compare them with existing SOTA methods. As in Tab.~\ref{nus_test}, MDRNet dramatically  improves CenterPoint~\cite{Yin2020centerpoint} to 65.2\% mAP and 70.5\% NDS. Moreover, without any testing augmentation, motorcycle and bicycle categories are remarkably increased to 73.1\% and 45.2\% AP, respectively. This is because the geometry of motorcycles and bicycles is more complex than other categories, and MDRNet is able to retain 3D geometric information more effectively than other methods.

For multi-modal settings, MDRNet-F with a simple fusion mechanism outperforms other methods with complex fusion strategies. This is attributable to the fact that SDR and MSR are able to reduce the loss of semantic and geometric information during dimensional collapse. With test-time augmentations~\cite{Yin2020centerpoint}, MDRNet-F$\;^{\dagger}$ further achieves 69.8\% mAP and 73.5\% NDS. Compared to Focals Conv~\cite{chen2022focal}, which also improves upon CenterPoint~\cite{Yin2020centerpoint}, the proposed MDRNet performs better both in LiDAR-only and multi-modal settings.
% \textbf{KITTI.}

\subsection{Ablation Studies}
\textbf{Improvements.} We evaluate the impact of our methods on existing SOTA detectors on the KITTI {\em val.} set and nuScenes {\em val.} set, respectively. On KITTI~\cite{geiger2013kitti}, we take PV-RCNN~\cite{shi2020pv} as a strong baseline. Compared to PV-RCNN, Tab.~\ref{kitti_val} shows that our method achieves appreciable improvement in the {\em pedestrian} category, boosting the AP$_{\textrm{3D}}$ from 54.49\% to 60.06\%. Tab.~\ref{nus_val} presents the comparison results on nuScenes~\cite{caesar2020nuscenes} {\em val.} set. Obviously, the proposed method significantly improves the performance of CenterPoint~\cite{Yin2020centerpoint}. It is worth noting that MDRNet with the voxel size set to 10 cm outperforms CenterPoint with the voxel size set to 7.5 cm.
% In Tab.~\ref{kitti_val}, we take PV-RCNN~\cite{shi2020pv} as a strong baseline on KITTI, compared to which our method achieves appreciable improvement over it, especially in the {\em pedestrian} category. 
% In Tab.~\ref{nus_val}, we take CenterPoint~\cite{Yin2020centerpoint} 

\begin{table}[h]    % htp
\centering
\caption{Improvements on PV-RCNN in AP$_{\textrm{3D}}$~(R40) of moderate difficulty on KITTI {\em val}.}
\label{kitti_val}
\begin{tabular}{l|ccc}
\hline
% \multirow{2}{*}{Methods} & \multicolumn{3}{c|}{AP$_{\textrm{3D}}$ (Mod.)} \\  
\multirow{1}{*}{Method} & Car & Pedestrian & Cyclist \\ \hline
% & Mod. & Mod. & Mod. \\
PV-RCNN & 84.36 & 54.49 & 70.38 \\
% Focals + PV-RCNN & \textbf{85.27} & 58.28 & 70.74 \\ 
PV-RCNN + ours & \textbf{84.85} & \textbf{60.06} & \textbf{70.96} \\
\hline
\end{tabular}
\end{table}

\begin{table}[h]
    \centering
    \caption{Improvements over CenterPoint on the nuScenes {\em val.} split.}
    \label{nus_val}
    \begin{tabular}{l|c|cc}
    \hline
    Method & Voxel Size  & mAP & NDS \\ \hline 
    CenterPoint & \multirow{2}{*}{10cm} & 56.38 & 64.79  \\
    CenterPoint + ours & & 59.81 & 67.09  \\ \hline  
    CenterPoint & \multirow{2}{*}{7.5cm} & 59.55 & 66.75  \\
    CenterPoint + ours & & \textbf{61.18} & \textbf{67.91}  \\
    % Ours-F & 70.9 & 67.1 \\
    \hline
    \end{tabular}
    
\end{table}

% \textbf{Impact of different components.} We next analyze the impact of the proposed two main contributions on 3D object detection. We use the backbone of PillarNet-18~\cite{shi2022pillarnet} as the baseline without adding the trick of IOU regression.
% \begin{table}[h]
%     \centering
%     \caption{Ablations on different modules of MDRNet.}
%     \label{ab_sdr_msr}
    
%     \begin{tabular}{cc|cc}
%     \hline
%         SDR & MSR & mAP & NDS \\ \hline
%             &     & 57.81 & 66.00   \\
%         \cmark &  & 59.06 & 66.49  \\
%          & \cmark &  60.41 & 67.72     \\
%         \cmark & \cmark & 61.18 & 67.91 \\ \hline     
%     \end{tabular}
% \end{table}

\textbf{Dimensionality reduction operations.} \label{sb_sdr} We conduct ablation experiments on the nuScenes~\cite{caesar2020nuscenes} {\em val.} set to explore the design of dimensionality reduction operations for grid-based 3D detectors. The ablations consist of two parts: various dimensionality reduction operations and the forms of spatial correlation distribution~(i.e. $w^{ReLU}, w^{Sigmoid}$ and $w^{Softmax}$). Tab.~\ref{ab_sdr} indicate that SDR-Softmax achieves the best performance among the four different dimensionality reduction operations,
% In Tab.~\ref{ab_sdr}, we compare our SDR with other dimensionality reduction operations, including mean/max pooling and sparse convolution. The results indicate that our SDR achieves the best performance, 
which should be attributed to the fact that Z-axis feature aggregation using spatial correlation distribution estimation is able to preserve more 3D geometric information. 
% In Tab.~\ref{ab_sdr_form}, $Softmax$ is superior to other forms in both mAP and NDS metrics. Therefore, it is adopted as the default setting.

% To further explore the effect of different different dimensionality reduction operations on grid-based point cloud detectors, 

\begin{table}[h]
\centering
\caption{Ablations on dimensionality reduction operations on the nuScenes {\em val} split.}
\label{ab_sdr}
\begin{tabular}{l|cc}
\hline
Operations & mAP  & NDS \\ \hline  %& Bicycle \\ \hline 
Mean Pooling & 60.11 & 67.36 \\ %& 46.5   \\
Max \ Pooling & 60.41 & 67.72 \\ %& 47.4   \\
Sparse Conv & 60.37 & 67.45 \\ \hline  %& 46.3 \\ \hline 
SDR-ReLU & 60.18 & 67.61 \\ %& 47.6 \\
SDR-Sigmoid & 60.37 & 67.50 \\ %& 46.7  \\
SDR-Softmax & \textbf{61.18} & \textbf{67.91} \\\hline  %& \textbf{49.0}  \\ \hline 
\end{tabular}
\end{table}

% \begin{table}[h]
% \centering
% \caption{Ablations on representations of spatial correlation distribution on the nuScenes {\em val} split.}
% \label{ab_sdr_form}
% \begin{tabular}{l|cc|c}
% \hline
% Normalization & mAP & NDS & Bicycle  \\ \hline
% ReLU & 60.18 & 67.61 & 47.6 \\
% Sigmoid & 60.37 & 67.50 & 46.7  \\
% Softmax & \textbf{61.18} & \textbf{67.91} & \textbf{49.0} \\ \hline 
% \end{tabular}
% \end{table}

% \textbf{Multi-level spatial residuals.} The purpose of multi-level spatial residuals~(MSR) is to retain multi-scale 3D information. 
% Tab.~\ref{ab_msr} demonstrates the results of using cross-dimensional connection~Sec.\ref{sec_msr} in different stages of the backbone network.
% The first row displays the results of using the backbone network with only BEV branch, whose first stage BEV features are obtained from voxel features by SDR. As the number of stages used increases, the performance is enhanced until all stages are involved. The last row is our proposed multi-level spatial residuals~Sec.\ref{sec_msr}. It can be noticed that multi-level spatial residuals are helpful to strengthen the performance of BEV branch, with mAP and NDS metrics improving by +2.25\% and +1.44\%, respectively.

\textbf{Ablations of SDR and MSR.} 
% The purpose of multi-level spatial residuals~(MSR) is to retain multi-scale 3D information. Tab.~\ref{ab_msr} demonstrates the results of using cross-dimensional connection~Sec.\ref{sec_msr} in different stages of the backbone network.
We perform ablations using the same experimental setup for the SDR module and the stages for using MSR. In Tab.~\ref{ab_msr}, the first row displays the results of using a simple pillar backbone network with pooling voxelization. The second row shows the results of adding the use of SDR to obtain the first stage BEV features and the following lines show the results of adding MSR to different stages.
% The first row displays the results of using the backbone network with only BEV branch, whose first stage BEV features are obtained from voxel features by SDR. 
As the number of stages used increases, the performance is enhanced until all stages are involved. It can be noticed that Multi-level Spatial Residuals are helpful to strengthen the performance of BEV branch, with mAP and NDS metrics improving by +2.25\% and +1.44\%, respectively.
% It should be noted that SDR is only used in the first stage and sparse convolution is used in the other stages as the spatial feature spaces in the subsequent stages are downsampled to a limited size and are not sufficient for sparsity adaptation, which is consistent with the conclusion of sparsity adaptation in Focals~cite{chen2022focal}. 

% We validate this on the nuScenes~\cite{caesar2020nuscenes} {\em val} set in Tab.~\ref{ab_msr}. The first row displays the results of only using BEV branch for backbone network. The second row demonstrates the results of adding a cross-dimensional connection~(\ref{sec_msr}) to stage 2. 
% The second row demonstrates the results of adding multi-level spatial residuals to BEV branch. As shown in Tab.~\ref{ab_msr}, multi-level spatial residuals are helpful to strengthen the performance of BEV branch, with mAP and NDS metrics improving by +2.25\% and +1.44\%, respectively.

% \begin{table}[h]
% \centering
% \caption{Ablations on multi-level spatial residuals.}
% \label{ab_msr}
% \begin{tabular}{c|c|c|cc}
% \toprule
% \multicolumn{3}{c|}{Stages}  & \multirow{2}{*}{mAP} & \multirow{2}{*}{NDS} %& \multirow{2}{*}{Bicycle}  
% \\ \cmidrule(l){1-3} 
% 2 & 3 & 4 & \\ %&\\
% \hline
% - & - & - & 58.83 & 66.47 \\ % & 45.0 \\
% \cmark & - & - & 59.58 & 67.02 \\ % & 45.9 \\
% \cmark & \cmark & - & 60.11 & 67.29 \\ % & 45.7 \\
% \cmark & \cmark & \cmark & \textbf{61.18} & \textbf{67.91} \\ \hline %& \textbf{49.0}  \\ \hline 
% \end{tabular}
% \end{table}

\begin{table}[h]
\centering
\caption{Ablations of SDR and MSR.}
\label{ab_msr}
\begin{tabular}{c|c|c|c|cc} % p{6mm}
\toprule 
\multirow{2}{*}{SDR} & \multicolumn{3}{c|}{Stages in MSR} & \multirow{2}{*}{mAP} & \multirow{2}{*}{NDS} %& \multirow{2}{*}{Bicycle} 
\\ \cmidrule{2-4}
& \multicolumn{1}{c|}{2} & \multicolumn{1}{c|}{3} & \multicolumn{1}{c|}{4} & \\ %&\\
\hline
\multicolumn{1}{c|}{-} & \multicolumn{1}{c|}{-} & \multicolumn{1}{c|}{-} &  \multicolumn{1}{c|}{-} & 57.81 & 66.00 \\
\multicolumn{1}{c|}{\cmark} & \multicolumn{1}{c|}{-} & \multicolumn{1}{c|}{-} & \multicolumn{1}{c|}{-} & 58.83 & 66.47 \\ % & 45.0 \\
\multicolumn{1}{c|}{\cmark} & \multicolumn{1}{c|}{\cmark} & \multicolumn{1}{c|}{-} & \multicolumn{1}{c|}{-} & 59.58 & 67.02 \\ % & 45.9 \\
\multicolumn{1}{c|}{\cmark} & \multicolumn{1}{c|}{\cmark} & \multicolumn{1}{c|}{\cmark} & \multicolumn{1}{c|}{-} & 60.11 & 67.29 \\ % & 45.7 \\
\multicolumn{1}{c|}{\cmark} & \multicolumn{1}{c|}{\cmark} & \multicolumn{1}{c|}{\cmark} & \multicolumn{1}{c|}{\cmark} & \textbf{61.18} & \textbf{67.91} \\ \hline %& \textbf{49.0}  \\ \hline 
\end{tabular}
\end{table}

% \begin{table}[]
%     \centering
%     \caption{Different combinations of the dimensionality reduction operations used in the first stage and MSR. $\circ$ indicates SDR and $*$ indicates sparse convolution}
%     \label{ab_combination}
%     \begin{tabular}{c|c|cc}
%     \toprule 
%     $1^{st}$ stage   &  MDR~($2^{nd},3^{rd},4^{th}$ stages) & mAP & NDS \\ \hline 
%       $*$ & $*$ & 60.37 & 67.45 \\ 
%       $\circ$  &  $\circ$ & 60.49 & 67.73 \\ 
%       $\circ$  & $*$ & 61.18 & 67.91 \\ \hline 
%     \end{tabular}
% \end{table}

\subsection{Runtime analysis}
The runtime of MDRNet and the baseline method CenterPoint~\cite{Yin2020centerpoint} is compared in Tab.\ref{timing} to evaluate the efficiency of our method. The experiments are conducted on an RTX 3090 GPU and all methods use the voxel size of (0.075m, 0.075m, 0.2m). On nuScenes~\cite{caesar2020nuscenes}, the inference speed of CenterPoint using MDRNet as backbone is almost the same as that of the original Centerpoint, yet the performance takes a substantial improvement. This result demonstrates that our proposed SDR~(Sec.~{\ref{sec_sdr}}) and MSR~(Sec.~{\ref{sec_msr}}) can boost the performance of existing 3D detectors without increasing the inference time.
% , and provide a better trade-off between runtime and accuracy.
\begin{table}[h]
    \centering
    \caption{Comparisons of inference time on the nuScenes}
    \label{timing}
    \begin{tabular}{c|c|cc}\hline
         Method & Runtime & mAP & NDS \\ \hline
         CenterPoint & 99ms & 59.55 & 66.75\\ \hline
         Ours & 100ms & 61.18 & 67.91 \\\hline
    \end{tabular}
    
\end{table}

\subsection{Qualitative results}
Qualitative results on the nuScenes {\em val.} set~\cite{caesar2020nuscenes} is shown in Fig.\ref{viz}. The first column displays the ground-truth boxes, and the second and third columns present the results for CenterPoint without and with our MDRNet, respectively. Compared to the original CenterPoint~\cite{Yin2020centerpoint}, MDRNet is able to dynamically concentrate on the valuable parts of the objects, effectively enhancing the perception capability. In Fig.\ref{viz}(c), the points inside the bounding boxes are colored according to the predicted spatial distribution. The closer the color is to red, the more valuable the area is considered by the network.
% , and the closer the color is to red indicates the greater the weight. 

\section{CONCLUSION}
% In this paper, we propose Spatial-aware Dimensionality Reduction~(SDR) and Multi-level Spatial Residuals~(MSR), which can better preserve 3D geometric information and reduce information loss during dimensional collapse. For the first time, we explore the effect of different dimensionality reduction operations on grid-based point cloud object detectors.
% Based on SDR and MSR, we design a novel point cloud backbone network called MDRNet, which enables existing state-of-art grid-based 3D detectors to achieve better performance on KITTI and nuScenes.
In this paper, we design a universal backbone network called MDRNet, which is able to be used with any grid-based point cloud detectors to enrich 3D geometry information. The proposed backbone is novel in two modules: Spatial-aware Dimensionality Reduction~(SDR) and Multi-level Spatial Residuals~(MSR). SDR performs adaptive feature aggregation along the height dimension by dynamically focusing the valuable parts of objects and MSR enriches the information of BEV features by multi-level 3D-BEV connections.  For the first time, we explore the effect of different dimensionality reduction operations on grid-based point cloud object detectors. Extensive experiments show that our MDRNet achieves SOTA performance on the large-scale nuScenes.


%%%%%%%%%%%%%%%%%%%%%%%%%%%%%%%%%%%%%%%%%%%%%%%%%%%%%%%%%%%%%%%%%%%%%%%%%%%%%%%%


%%%%%%%%%%%%%%%%%%%%%%%%%%%%%%%%%%%%%%%%%%%%%%%%%%%%%%%%%%%%%%%%%%%%%%%%%%%%%%%%
% \section*{ACKNOWLEDGMENT}
% This work was supported by the Tencent YouTu Lab and Tsinghua University.

%%%%%%%%%%%%%%%%%%%%%%%%%%%%%%%%%%%%%%%%%%%%%%%%%%%%%%%%%%%%%%%%%%%%%%%%%%%%%%%%
\clearpage
\bibliography{root.bib}

\end{document}